\documentclass{article} 
\usepackage{nips12submit_e,times}
\usepackage{algorithm}
\usepackage{algorithmic}
\usepackage{amsmath, amssymb, amsfonts, mathrsfs}
\usepackage{subfigure}
\usepackage{graphicx}
\usepackage{natbib}

\title{Sparse Penalty in Deep Belief Networks: Using the Mixed Norm Constraint}

\author{
Xanadu C. Halkias \\
DYNI, LSIS,
Universit\`{e} du Sud, \\
Avenue de l'Universit\'{e} - BP20132, 83957 LA GARDE CEDEX - FRANCE \\
\texttt{xanadu.halkias@univ-tln.fr} 
\And
S\'{e}bastien Paris \\
DYNI, LSIS CNRS UMR 7296, Aix-Marseille University Domaine universitaire de Saint J\'{e}r\^{o}me  \\
Avenue Escadrille Normandie Niemen,13397 MARSEILLE Cedex 20, FRANCE \\
\texttt{sebastien.paris@lsis.org} 
\And
Herv\'{e} Glotin\\
DYNI, LSIS CNRS UMR 7296, Universit\'{e} Sud Toulon-Var, Institut Universitaire de France \\
Avenue de l'Universit\'{e} - BP20132, 83957 LA GARDE CEDEX - FRANCE \\
\texttt{glotin@univ-tln.fr}
}

\nipsfinalcopy 

\begin{document}

\maketitle
\newcommand{\g}[1]{\mbox{\boldmath{$#1$}}}
\begin{abstract}
Deep Belief Networks (DBN) have been successfully applied on popular machine learning tasks. Specifically, when applied on hand-written digit recognition, DBNs have achieved approximate accuracy rates of $98.8\%$. In an effort to optimize the data representation achieved by the DBN and maximize their descriptive power, recent advances have focused on inducing sparse constraints at each layer of the DBN. In this paper we present a theoretical approach for sparse constraints in the DBN using the mixed norm for both non-overlapping and overlapping groups. We explore how these constraints affect the classification accuracy for digit recognition in three different datasets (MNIST, USPS, RIMES) and provide initial estimations of their usefulness by altering different parameters such as the group size and overlap percentage.
\end{abstract}

\section{Introduction}

\label{intro}

Restricted Boltzmann Machines (RBMs) are Energy Based Models (EBMs) that have been extensively used for a diverse set of machine learning applications mainly due to their generative and unsupervised learning framework. These applications range from image scene recognition and generation~\cite{Hinton2006b}, video-sequence recognition~\cite{Hinton2007} and dimensionality reduction~\cite{Hinton2006}.

An equally important aspect of RBMs is that they serve as the building blocks of DBNs~\cite{BengioNIPS2006}. Their use as such has been favored in the machine learning community due to the conditional independence between the hidden units in the RBM that allows for the efficient and computationally tractable implementation of deep architectures.

In recent years, sparsity has become an important requirement in both shallow [add cite] and deep architectures. Although primarily used in statistics for optimization tasks in order to overcome the curse of dimensionality in various applications, it also serves as a way to emulate biologically plausible models of the human visual cortex, where it has been shown that sparsity is an integral process in the hierarchical processing of visual information~\cite{LeeNIPS2007,Field1997,farabetpami}.

Moreover, an added benefit of using sparse constraints in the form of mixed norm regularizers in deep architectures is that they can alleviate their restrictive nature by allowing implicit interactions between the hidden units in the RBMs. Mixed norm regularizers such as $l_{1,2}$ have been extensively used in statistics and machine learning~\cite{OptMachineLearning}. In this paper we provide initial results when inducing sparse constraints by using a mixed norm regularizer on the activation probabilities of the RBMs. The mixed norm is applied on both non-overlapping and overlapping groups. We also show that this regularizer can be used to train DBNs, and offer results for the task of digit recognition using several datasets.

\section{Restricted Boltzmann Machines}
\label{RBM}

An RBM is a type of two layer neural network comprised of a visible layer that represents the observed data $x$ and a hidden layer that represents the hidden variables $h$. The addition of these hidden units allows the model an increased capacity in expressing the underlying distribution of the observed data.

RBMs are energy based models and as such they define a probability distribution through an energy function as seen in Eq.~\ref{energy}
\begin{equation}
     p(\textbf{x},\textbf{h}) = \frac{e^{-E(\textbf{x},\textbf{h})}}{Z}
\label{energy}
\end{equation}
Where $Z$, provided in Eq.~\ref{partition}, is called the partition function and is a normalizing factor ensuring that Eq.~\ref{energy} is a probability.
\begin{equation}
     Z = \sum_{\textbf{x},\textbf{h}}e^{-E(\textbf{x},\textbf{h})}
\label{partition}
\end{equation}
In the case of an RBM the energy function $E(\textbf{x},\textbf{h})$ is defined in Eq.~\ref{RBMen}:
\begin{equation}
      E_{\theta}(\textbf{x},\textbf{h}) = -\sum_{i=1}^{I}\sum_{j=1}^{J}x_ih_jw_{ij}-\sum_{i=1}^{I}b_ix_i-\sum_{j=1}^{J}a_jh_j,
\label{RBMen}
\end{equation}

$\textbf{b}$ are the visible unit biases and $\g{a}$ are the hidden unit biases.

In the common case where we are using stochastic binary units for both visible and hidden units, then the conditional probabilities of activation are obtained by:
\begin{equation}
   \begin{array}{l}
      p(x_i = 1|\textbf{h}) = \sigma(b_i+\sum_j h_jw_{ij})\\
      p(h_j = 1|\textbf{x}) = \sigma(a_j+\sum_i x_iw_{ij}),
   \end{array}
\label{condsigm}
\end{equation}
where $\sigma$ is the sigmoid function and
\begin{equation}
\sigma(f(x)) \triangleq \frac{1}{1+e^{-f(x)}}.
\end{equation}

Since an RBM does not allow for connections amongst hidden units or amongst visible units we can easily obtain Eq.~\ref{cond}.
\begin{equation}
   \begin{array}{l}
      p(\textbf{x}|\textbf{h}) = \prod\limits_{i}p(x_i|\textbf{h})\\
      p(\textbf{h}|\textbf{x}) = \prod\limits_{j}p(h_j|\textbf{x})
   \end{array}
\label{cond}
\end{equation}

Intuitively, the observed data, $\textbf{x}$ will be modeled by those hidden units, $\textbf{h}$ that are expressed with a high conditional probability $p(h_j|\textbf{x})$. The goal of adding sparse constraints to the network is to allow for the salient activation of the hidden units based on the differences of the observed data. As a result, we can achieve an initial clustering of the observed data that will increase the discriminative power of the model.

\subsection{Training an RBM}

RBMs are energy based, generative models that are trained to model the marginal probability $p(\boldsymbol{x})$ of the observed data where:
\begin{equation}
p(\textbf{x}) = \sum_{\textbf{h}\in \{0,1\}^J} p(\textbf{x},\textbf{h}).
\end{equation}

In general, energy based models can be learnt by performing gradient descent on the negative log-likelihood of the observed data. Specifically, to learn the parameters of the network we need to compute the gradient provided in Eq.~\ref{NLL} given the observed (training) data $\boldsymbol{x^l}$.
\begin{equation}
    -\frac{\partial logp(\textbf{x})}{\partial\boldsymbol{\theta}} = \langle\frac{\partial E_{\theta}(\boldsymbol{x^l},\textbf{h})}{\partial\boldsymbol{\theta}}\rangle_{\boldsymbol{h}} - \langle\frac{\partial E_{\theta}(\textbf{x},\textbf{h})}{\partial\boldsymbol{\theta}}\rangle_{\boldsymbol{x},\boldsymbol{h}},
\label{NLL}
\end{equation}

where $\langle \cdot \rangle_n$ denotes the expectation with respect to $n$. As evident in Eq.\ref{NLL}, the gradient has two phases. The positive phase which tries to lower the energy of the training data $\boldsymbol{x^l}$ and the negative phase which tries to increase the energy of all $x$ in the model. \\
Assessing the energy on all the data can be an intractable task given the size of the network and the number of possible configurations. In order to obtain an approximation Hinton (2006) successfully proposed the use of Contrastive Divergence (CD). This allows us to sample an approximation of the expectation over $(\textbf{x},\textbf{h})$ using Gibbs sampling at only $k$ steps. Empirically, it has been shown that setting $k = 1$ will provide an adequate approximation although it will not follow the theoretical gradient~\cite{BengioNIPS2006}.

Applying CD on Eq.~\ref{NLL} we can obtain the following update equations for the parameters of the network.
\begin{equation}
\begin{array}{l}
  \Delta \g{w}_{\cdot j} = \frac{1}{L}\sum\limits_{l=1}^{L}\g{x^l}p(h_j=1|\g{x^l}) - \g{\widetilde{x}^l}p(h_j=1|\g{\widetilde{x}^l})
\end{array}
\label{weightrbm}
\end{equation}
\begin{equation}
\begin{array}{l}
  \Delta b_i = \frac{1}{L}\sum\limits_{l=1}^{L}p(x_i^l=1|\g{h}) - p(\widetilde{x}_i^l=1|\g{h})
\end{array}
\label{bivisrbm}
\end{equation}
\begin{equation}
\begin{array}{l}
\Delta a_j = \frac{1}{L}\sum\limits_{l=1}^{L}p(h_j=1|\g{x^l}) - p(h_j = 1|\g{\widetilde{x}^l}),
\label{bihidrbm}
\end{array}
\end{equation}
where the $(\widetilde{\cdot})$ defines the generated distributions obtained by the CD.

In the next section, we introduce a general version of sparse constraints in the learning phase of the RBM through the use of the mixed norm in an effort to control the activation probabilities of the hidden units.
\section{Mixed Norm RBMs}
\label{SGLRBM}

Several attempts in inducing sparse constraints in the RBM by~\cite{LeeNIPS2007, RanzatoNIPS2007} have been successful in increasing the discriminative power of the models. Examples of these sparse constraints range from weight decay~\cite{Hinton_guide} to modified norm penalties ~\cite{RanzatoNIPS2006}. In this paper we focus on the generalized penalty of the mixed norm ($l_{1,2}$), but will also provide a theoretical and practical implementation for the use of overlapping groups. We will refer to this generalized penalty applied to the expectations of the activation probabilities as the Mixed Norm RBM (MNRBM).

As mentioned before, learning an RBM consists of performing gradient descent on the negative log-likelihood. We can thus define the cost function $L$ to be minimized as $L = -logp(\boldsymbol{x})$. When applying the mixed norm regularizer the cost function takes the general form of Eq.~\ref{costsparse}.
\begin{equation}
\begin{array}{l}
  L = -logp(\textbf{x}) + \lambda(\|p(\textbf{h}=1|\textbf{x})\|_{1,2})
\end{array}
\label{costsparse}
\end{equation}
Where $\lambda$ is a regularizer constant.
The second term of Eq.~\ref{costsparse} defines the mixed norm penalty on the expectations of the hidden unit activation probabilities. In order to apply the mixed norm we assume that the hidden units are divided into groups. These groups can be non-overlapping or overlapping. As a result, we are able to penalize a whole group and not just individual hidden units.\\

{\bf MNRBM with non-overlapping groups:} Given an RBM with $J$ hidden units we define a partition of the hidden units into groups $P_m$ where $m = 1,2,...M$. The groups are non-overlapping and of equal size to alleviate computational issues. The mixed norm penalty for a data sample $\g{x^l}$ is defined in Eq.~\ref{gl}.
\begin{equation}
  \begin{array}{ll}
  \|p(\textbf{h}=1|\g{x^l})\|_{1,2} & = \sum\limits_{m = 1}^{M}\|p(\g{P_m}|\g{x^l})\|_2\\
  & = \sum\limits_{m=1}^{M}\sqrt{\sum\limits_{k \in P_m} p(h_k = 1|\g{x^l})^2}
  \end{array}
\label{gl}
\end{equation}
In practice, the desire behind the application of the mixed norm penalty is to set groups of the hidden units to zero when representing the observed data by forcing their activation probabilities to zero. As a result, given an observed data sample only a small number of groups of hidden units will be activated, leading to its sparse representation.

{\bf MNRBM with overlapping groups:} Given an RBM with $J$ hidden units we define a partition of the hidden units into groups $P_m$ where $m = 1,2,...M$. The groups are overlapping and of equal size. Depending on the percentage of overlap, $a$ we will obtain a new set of groups $P'_k$ where $k = 1,2,...K$.\\
We can then define a set of augmented hidden units $J' = \{h'\in J: \forall P_k, P_k \bigcup P_m = P_m, J' \supset J\}$. Subsequently, given an RBM with $J'$ hidden units, we can then consider that the set $P'_k$, defines non-overlapping, equally sized groups~\cite{jacobicml2009}. The mixed norm penalty for a data sample $\g{x^l}$ is defined in a similar way as in Eq.~\ref{gl}.
\begin{equation}
  \begin{array}{ll}
  \|p(\g{h'}=1|\g{x^l})\|_{1,2} & = \sum\limits_{k = 1}^{K}\|p(\g{h'}_m|\g{x^l})\|_2\\
  & = \sum\limits_{k=1}^{K}\sqrt{\sum\limits_{k \in P'_k} p(h'_k = 1|\g{x^l})^2}
  \end{array}
\label{gl}
\end{equation}

\subsection{Training the Mixed Norm RBM} 
In order to train the MNRBM with non-overlapping groups and obtain the model parameters $\g{\theta}$ we need to minimize the cost function presented in Eq.~\ref{costsparse}. This can be achieved by performing a coordinate descent once we have obtained the gradients of the regularizers.

The gradient of the mixed norm penalty for the weights, $W$ is as follows:
\begin{equation}
      \begin{array}{l}
      \frac{\partial }{\partial \g{w}_{\cdot j}}(\|p(\textbf{h}=1|\g{x^l})\|_{1,2})= \\
      = \frac{1}{2}\cdot \frac{1}{\sqrt{\sum\limits_{k \in P_m} p(h_k = 1|\g{x^l})^2}}\cdot 2 \cdot p(h_k=1|\g{x^l})\cdot \frac{\partial p(h_k=1|\g{x^l}) }{\partial \g{w}_{\cdot j}}\\
      = \frac{p(h_k=1|\g{x^l})}{\|p(\g{h}_m|\g{x^l})\|_2}\cdot \frac{\partial p(\g{h}_k = 1|\g{x^l})}{\partial w_{\cdot j}}\\
      = \frac{p(h_k=1|\g{x^l})}{\|p(\g{h}_m|\g{x^l})\|_2}\cdot p(h_k=1|\g{x^l})[1-p(h_k=1|\g{x^l})]\cdot \g{x^l}\\
      = \frac{p(h_k=1|\g{x^l})^2}{\|p(\g{h}_m|\g{x^l})\|_2}\cdot p(h_k=0|\g{x^l})\cdot \g{x^l}.
      \end{array}
\label{gradgl}
\end{equation}
When applied on the expectations of the activation probabilities the mixed norm penalty will follow their trend while forcing the groups that include members with low activation probabilities towards zero. The $l_2$ norm in the denominator ensures that the groups with low activations will be pushed further closer to zero.

Given the gradients of the penalties the update equations for the MNRBM are presented bellow:
\begin{equation}
  \begin{array}{l}
  \Delta \g{w}_{\cdot j} =\\
  \frac{1}{L}\sum \limits_{l=1}^{L}[(p(h_j = 1|\g{x^l})+\lambda\frac{p(h_j=1|\g{x^l})p(h_j = 0|\g{x^l})}{\sqrt{\sum\limits_{}p(h_m=1|\g{x^l})^2}})\cdot \g{x^l}-p(h_j = 1|\g{\widetilde{x}^l})\g{\widetilde{x}^l}] \\
  \end{array}
\label{weightsglrbm}
\end{equation}
\begin{equation}
\begin{array}{l}
  \Delta a_{j} = \\
  \frac{1}{L}\sum \limits_{l=1}^{L}[(p(h_j = 1|\g{x^l})+\lambda\frac{p(h_j=1|\g{x^l})p(h_j = 0|\g{x^l})}{\sqrt{\sum\limits_{}p(h_m=1|\g{x^l})^2}}-p(h_j = 1|\g{\widetilde{x}^l})] \\
\end{array}
\label{bihidsglrbm}
\end{equation}
 The detailed steps for training the MNRBM are depicted in Algorithm~\ref{alg:SGLRBMLearn}.

\begin{algorithm}[tb]
   \caption{Mixed Norm RBM learning algorithm}
   \label{alg:SGLRBMLearn}
   $1$. Update the parameters $\g{\theta}$ using CD and Eq.~\ref{weightrbm}-~\ref{bihidrbm}\\
   $2$. Update the parameters again using the gradient of the regularizations as in Eq.~\ref{weightsglrbm}-~\ref{bihidsglrbm}\\
   $3$. Repeat steps $1$, $2$ until convergence
\end{algorithm}

The general penalty of Eq.~\ref{costsparse} allows us through the manipulation of the constant regularizer, $\lambda$, the group size and percentage of overlap to obtain different types of architectures. In this case, the sparsity is induced at the group level of the hidden units whereby the observed data is represented by a small number of groups of hidden units. The $\lambda$ constant is empirically determined~\cite{Hinton_guide} based on the task at hand.

\begin{figure*}
\centering
\subfigure[Mixed norm RBM non-overlapping groups (group size = 20)]{\includegraphics[width = 60mm]{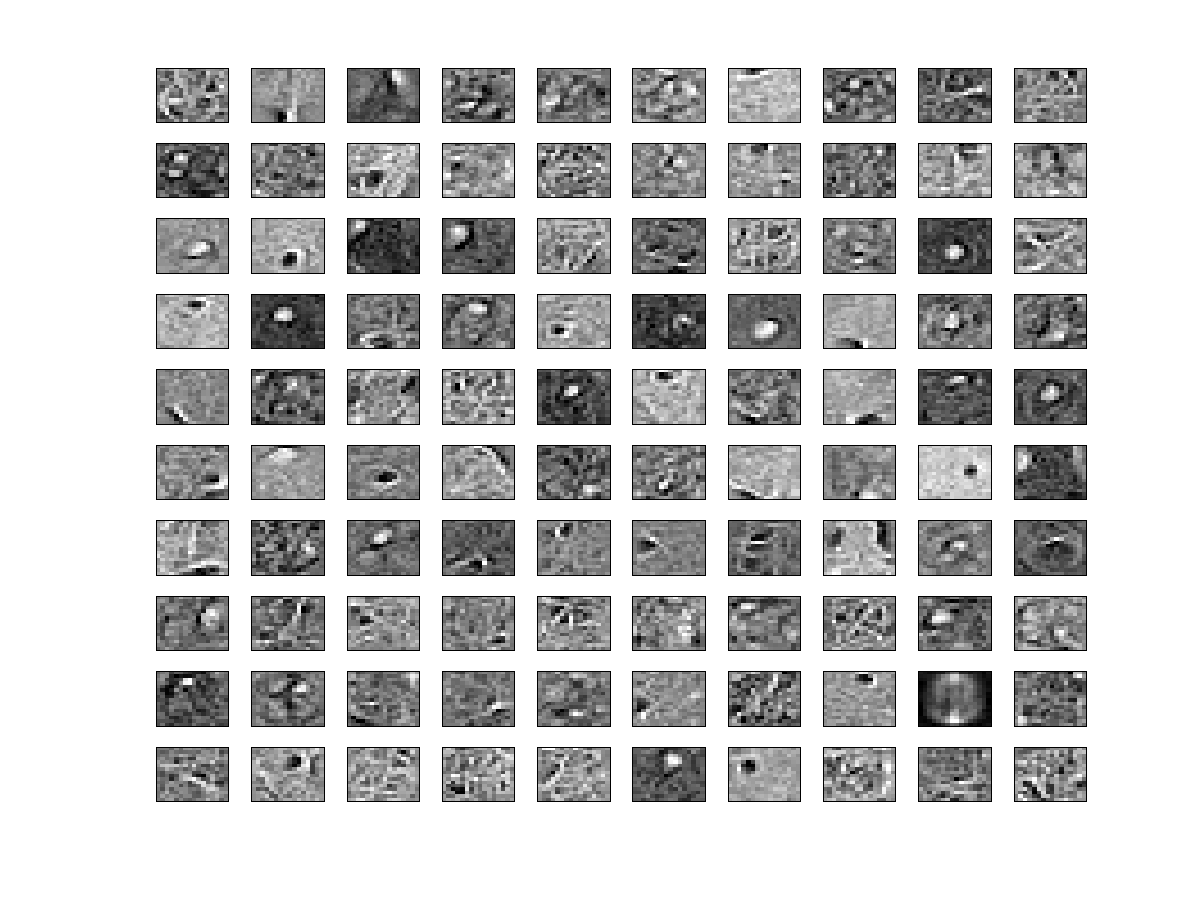}}
\subfigure[Mixed norm RBM overlapping groups (group size = 50, overlap = 20\%)]{\includegraphics[width = 60mm]{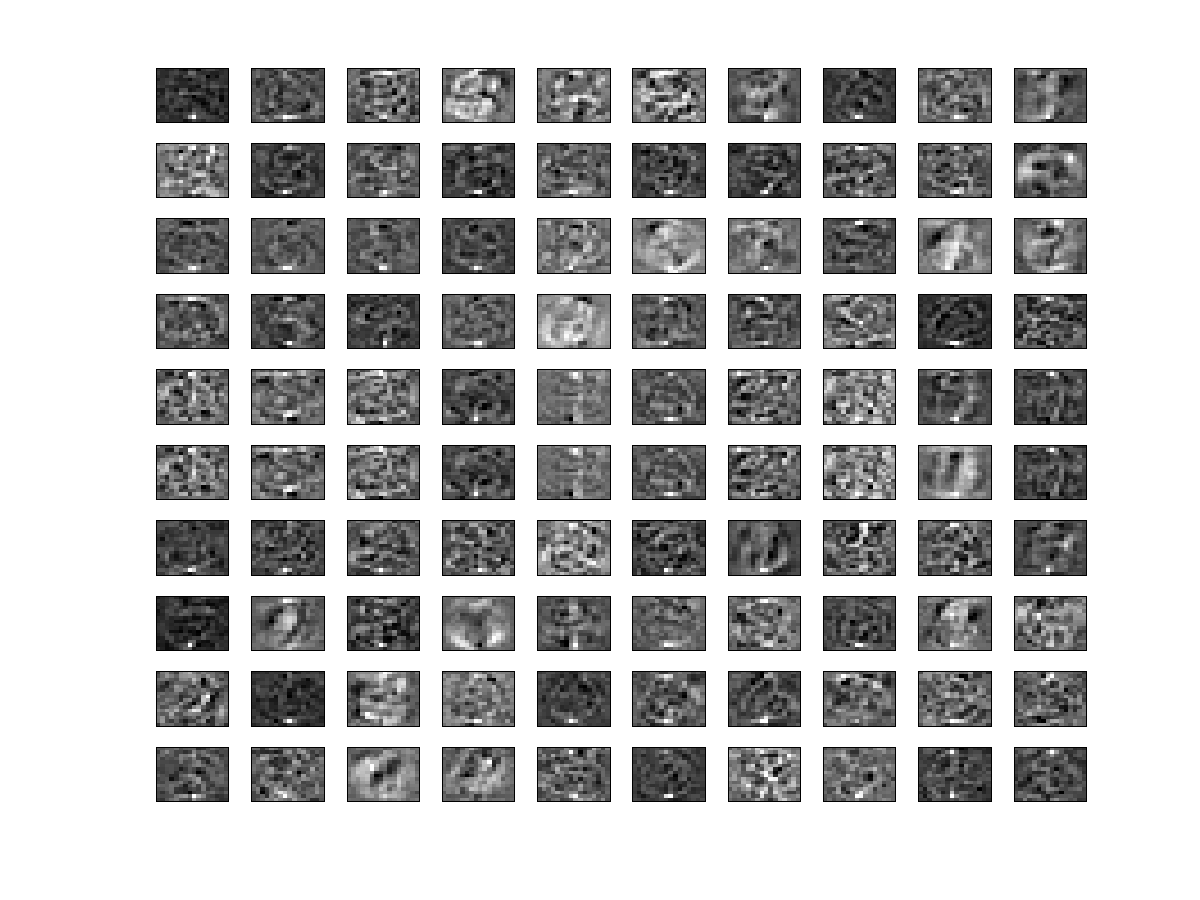}}
\caption{Sample learned weights $W$ for the mixed norm RBM using the USPS data set}
\label{fig:mn_weights}
\end{figure*}

\begin{figure*}
\centering
\subfigure[Mixed norm RBM non-overlapping groups (group size = 20)]{\includegraphics[width = 60mm]{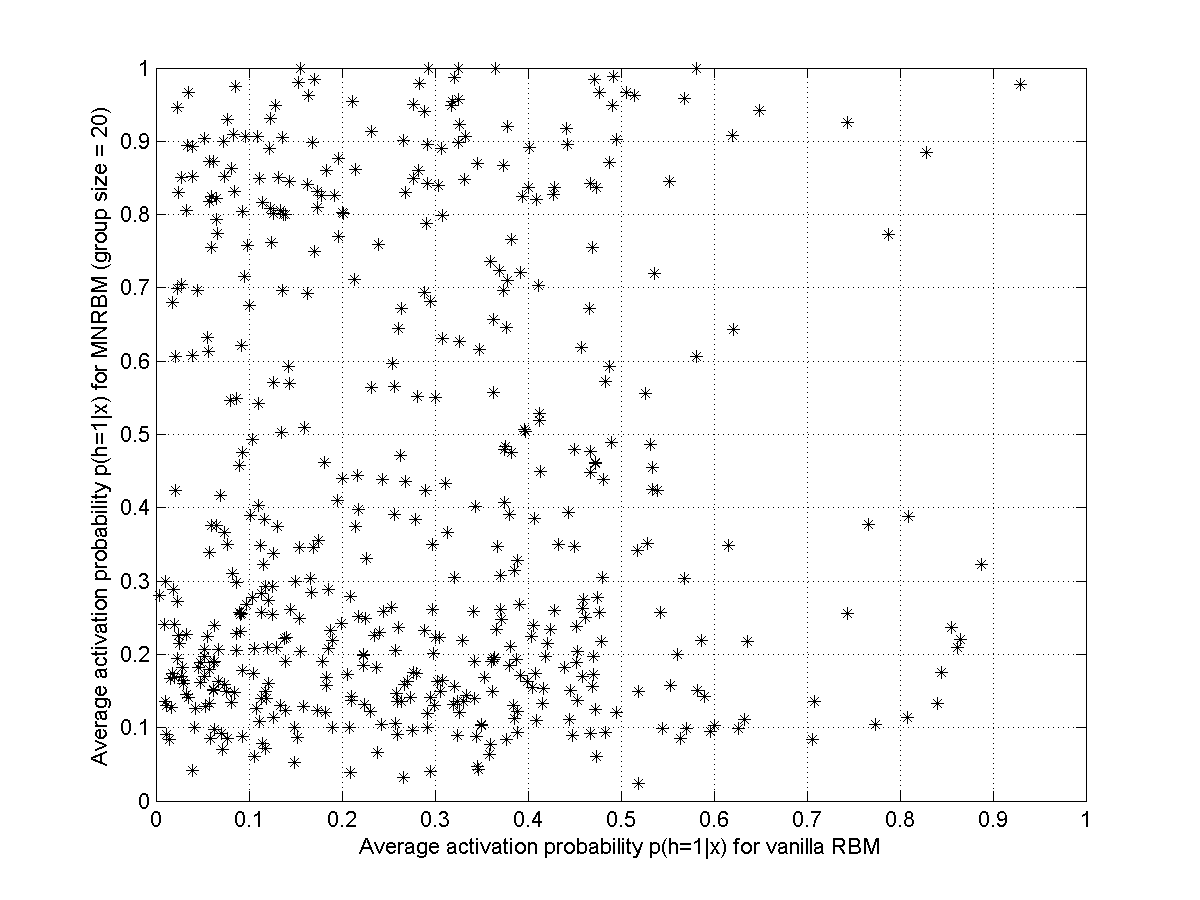}}
\subfigure[Mixed norm RBM overlapping groups (group size = 50, overlap = 20\%)]{\includegraphics[width = 60mm]{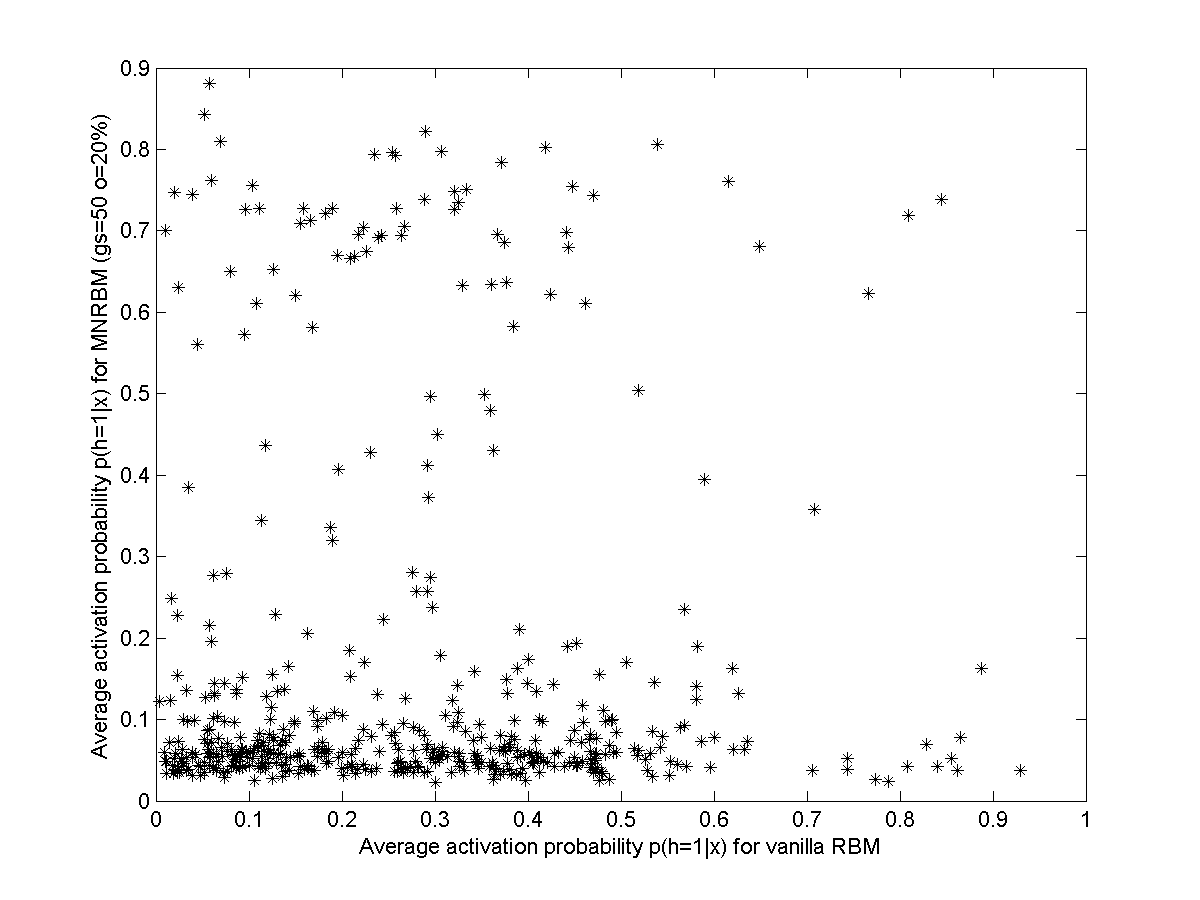}}
\caption{Average of hidden unit activation probabilities for the mixed norm RBM using a batch of the USPS data set. Y-axis: Hidden unit activation probabilities for mixed norm RBM. X-axis: Hidden unit activations for vanilla RBM}
\label{fig:mn_hact}
\end{figure*}

Fig.~\ref{fig:mn_weights} shows sample weights for the mixed norm RBM when using $\lambda = 0.1$ for both non-overlapping and overlapping groups. Fig~\ref{fig:mn_hact} provides the average probability activations for the hidden units given a batch of the USPS training data. As seen in the figure, the activation probabilities of the hidden units appear to be more towards the left-hand side of the figure which is the desired effect. However, there appears to be a bimodality whereby a large proportion of the activation probabilities is set to a high value for the non-overlapping groups MNRBM. This may be attributed to the choice and size of groups when applying the mixed norm penalty. Given that the activation probabilities are pushed towards high values one can expect that such a process may have an adverse result for classification tasks since the hidden units will over-represent the observed data. However, in the case of the overlapping groups most of the activations are pushed towards zero. Although, that is the goal of adding sparse constraints is to force the activations to zero, in this case we may be dealing with a biased system that actually under represents the data distribution.

\subsection{Data}
\begin{figure}[ht!]
\begin{center}
\centerline{\includegraphics[width=70mm]{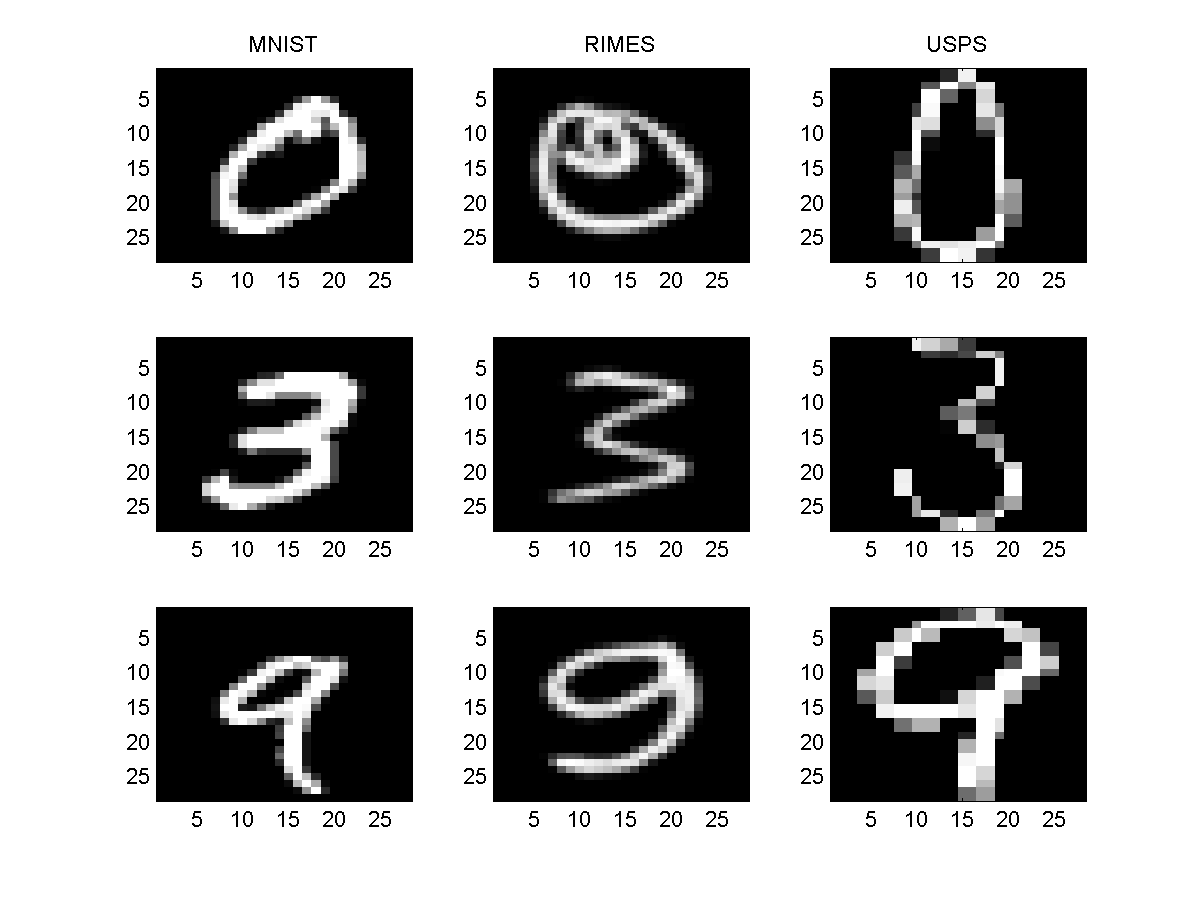}}
\caption{Examples of images from the three datasets MNIST (left), RIMES (center) and USPS (right)}
\label{data}
\end{center}
\vskip -0.2in
\end{figure}

We have used three different data sets in order to train and test the network.
\begin{itemize}
\item MNIST is a popular data set in the community for hand-written digit recognition and is comprised of $70000$, $28\times 28$ images ($60000$ train - $10000$ test). It is publicly available at \emph{yann.lecun.com/exdb/mnist}.
\item The RIMES data set which was created by asking volunteers to write hand written letters for different scenarios. In this paper we used the digit set of the data base. In total the set we used was comprised of $37200$ images of different sizes ($29800$ train - $7400$ test). Further information can be obtained at \emph{www.rimes-database.fr}.
\item The USPS digit data set that we used is comprised of $9280$ ($7280$ train - $2000$ test), $16\times 16$ images. The extracted images were scanned from mail in working U.S. Post Offices~\cite{USPS}.
\end{itemize}
In order to achieve the cross-training and testing all images were resized to have the same size as the MNIST dataset ($28\times 28$) given its extensive use in this task. All images were also checked to ensure that orientations/translations  were uniform across the data sets. No other pre-processing was employed. Example images from the three datasets can be seen in figure~\ref{data}

\subsection{Experimental Results: Pre-training DBNs with MNBMs}

RBMs became increasingly popular when Hinton and Salakhudinov~\cite{Hinton2006}~\cite{Hinton2006b} used them as building blocks for creating and pre-training efficient DBNs. The proposed MNRBMs can be utilized in the same manner to initialize DBNs and obtain a sparse and computationally efficient representation of the observed data.

In order to offer a comparative view between the different architectures we used Hinton's model for digit recognition, but we substituted the vanilla RBM with the proposed MNRBM. We pre-trained a $500-500-2000$ DBN and tested it on three different data sets, MNIST, RIMES and USPS.

Continuing, to obtain classification error rates we added $10$ softmax layers to get the posterior probabilities for the different classes. The network was fine-tuned using conjugate gradient as described in~\cite{Hinton2006}. The constant regularizer was empirically set to $\lambda = 0.1$ for all different models~\cite{Hinton_guide}. Continuing, for the mixed norm architecture with non-overlapping groups we used different group sizes for the hidden units, $5$, $10$, $20$ and $100$ respectively. In the case of overlapping groups we used group sizes of $20$, and $50$ with $a = \{20\%, 50\%\}$. Results on the classification accuracy and the computational cost of the models can be seen in Table~\ref{DBN-table} and Table~\ref{CPU-table} respectively. All experiments were performed on a $24$ core server (AMD Opteron processor $8435$) with a core CPU of $2.6$GHz and a cache of $512$KB.

From Table~\ref{DBN-table} we can infer that our proposed mixed norm penalty can offer the flexibility of creating architectures that will be able to matchthe classification accuracy of the models depending on the underlying distributions. It appears that for the task of hand-written digit recognition the distribution of the observed data favors the use of larger non-overlapping group sizes for the mixed norm architectures.

\begin{figure*}
\centering
\subfigure[Average Probability density function for activation probabilities, $p(\g{h=1}|\g{x})$]{\includegraphics[width = 60mm]{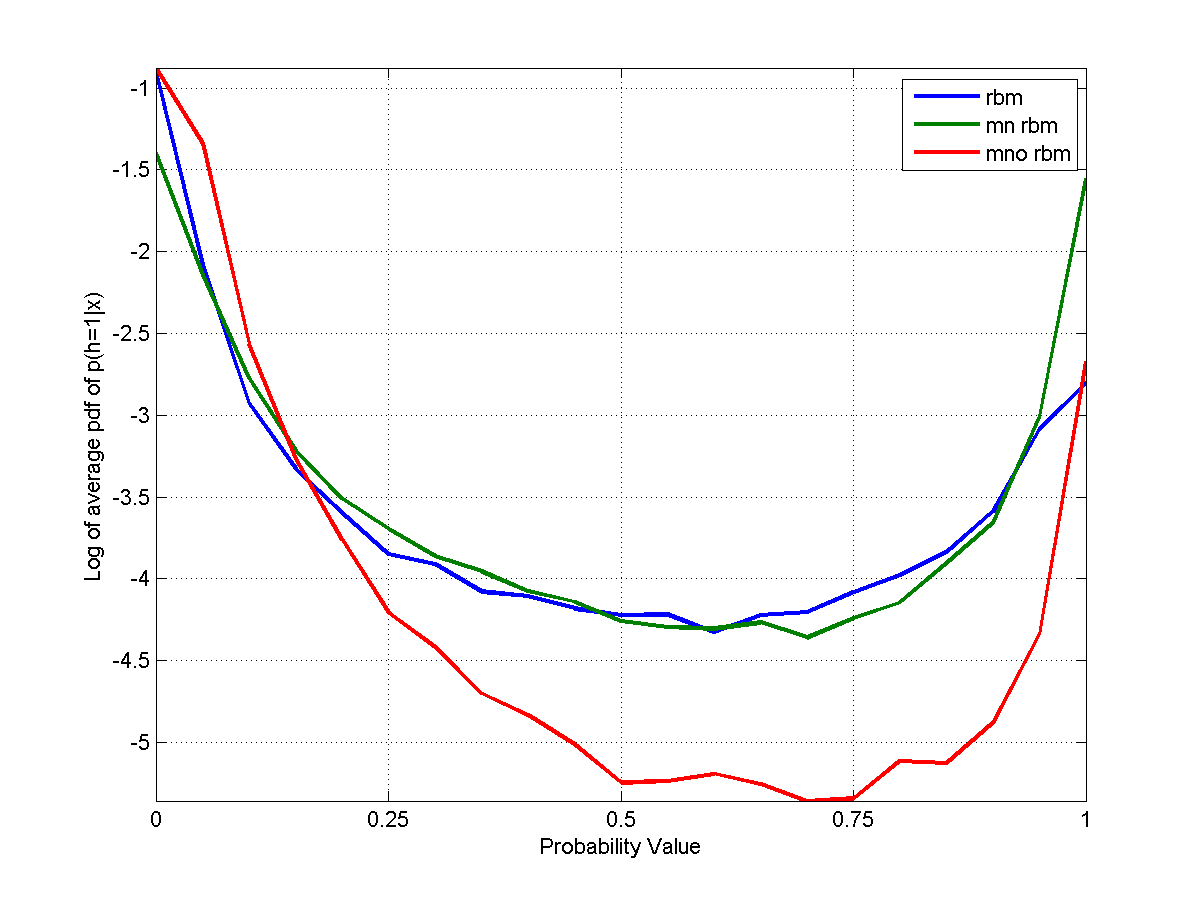}}
\subfigure[Classification Accuracy for the different penalties using the USPS data set]{\includegraphics[width = 60mm]{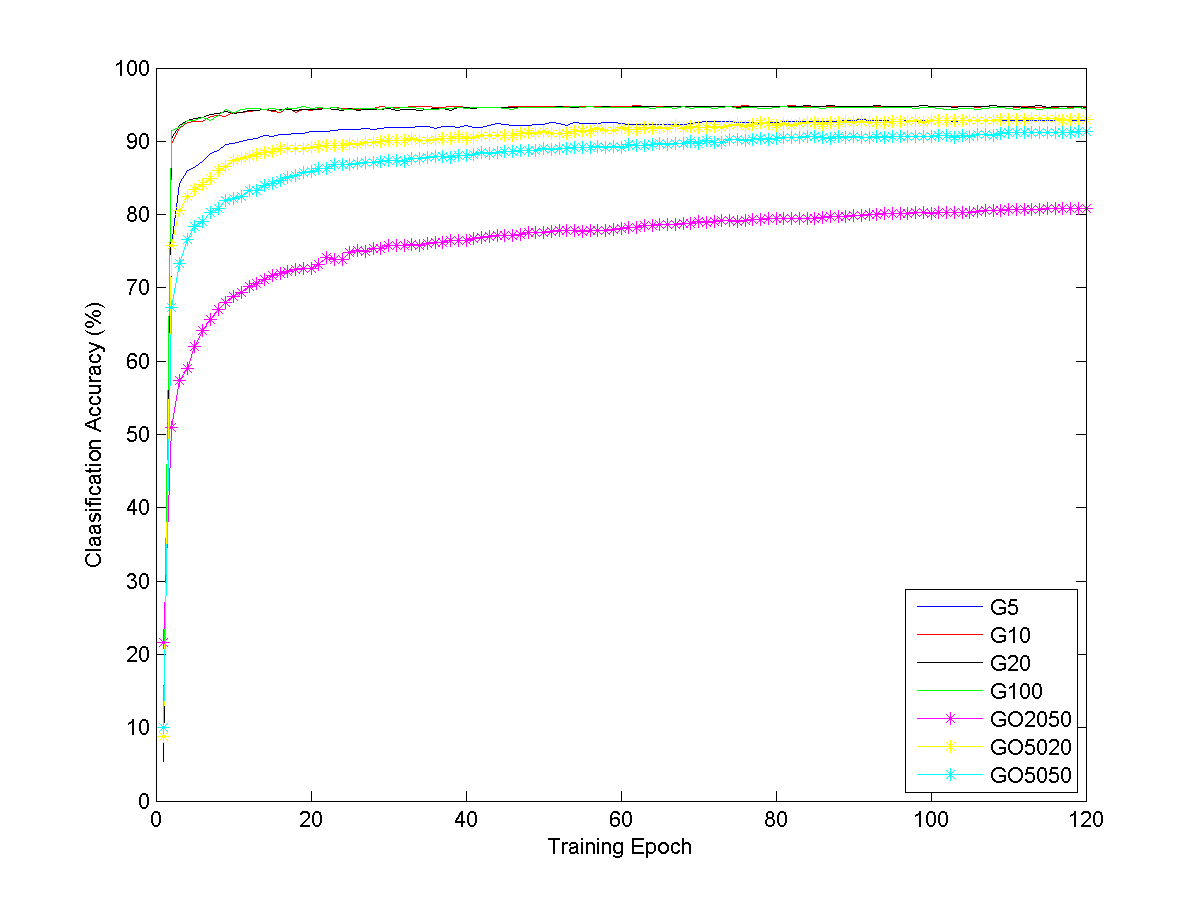}}
\caption{Average log pdf for activation probabilities for the vanilla RBM and mixed norm RBM using a batch of the USPS data set and classification accuracy for the USPS data set using the different architectures }
\label{fig:prob_pdf}
\end{figure*}

In order to get a better understanding of the impact of the different sparse constraints and architectures, Figure~\ref{fig:prob_pdf} depicts the average probability density functions of the expectations of the activation probabilities for the MNIST training data.

It is interesting to note that the proposed architectures that utilize the mixed norm penalty (MNDBN)with the overlapping groups tend to aggressively push their activation probabilities to zero. However, these architectures also tend to offer lower accuracy rates which can be attributed to an inability of the models to concisely capture the underlying data. A possible way for exploring this phenomenon further may be to constrain the penalty of the expectations as seen in~\cite{RanzatoNIPS2007}.

\begin{table}[t]
\caption{Classification accuracies for the different architectures
based on the general sparse penalty.}
\label{DBN-table}
\vskip 0.15in
\begin{center}
\begin{small}
\begin{sc}
\begin{tabular}{lcccr}
\hline
Architecture & MNIST & RIMES & UPS \\
\hline
DBN    & \g{98.83$\%$}& 99.30$\%$& \g{94.85$\%$} \\
MN DBN (5)& 97.28$\%$& 99.24$\%$&92.90$\%$\\
MN DBN (10)& \g{98.83$\%$}& 99.33$\%$&94.70$\%$ \\
MN DBN (20)& 98.77$\%$& 99.38$\%$ &94.65$\%$  \\
MN DBN (100)& 98.80$\%$& \g{99.40$\%$}&94.35$\%$  \\
MN w/O DBN (20/20\%)& 95.10$\%$ & 95.70$\%$ & 85.05$\%$\\
MN w/O DBN (20/50\%)& 93.50$\%$& 93.62$\%$& 80.90$\%$ \\
MN w/O DBN (50/20\%)& 96.50$\%$& 97.60$\%$& 92.95$\%$ \\
MN w/O DBN (50/50\%)& 95.84$\%$& 96.27$\%$& 91.35$\%$ 
\end{tabular}
\end{sc}
\end{small}
\end{center}
\vskip -0.1in
\end{table}
\begin{table}[th]
\caption{CPU times for the different architectures
based on the general sparse penalty.}
\label{CPU-table}
\vskip 0.15in
\begin{center}
\begin{small}
\begin{sc}
\begin{tabular}{lcccr}
\hline
Architecture & MNIST & RIMES & UPS \\
\hline
DBN    & 167.90h & $>$60h & 31.15h \\
MN DBN (5)& 62.14h& 33.70h& 8.62h\\
MN DBN (10)& 66.10h& 40.70h& 10.00h \\
MN DBN (20)& 70.10h& 69.80h& 12.75h \\
MN DBN (100)& 71.50h& 85.80h& 15.85h    \\
MN w/O DBN (20/20\%) & $>$60h& 39.27h& 10.40h\\
MN w/O DBN (20/50\%) & $>$60h& $>$45h& 22.90h \\
MN w/O DBN (50/20\%) & $>$60h& 35.60h& 9.56h \\
MN w/O DBN (50/50\%) & $>$70h& $>$45h& 24.00h 
\end{tabular}
\end{sc}
\end{small}
\end{center}
\end{table}

\subsection{Conclusions}
In this work we provided some first insights for the use of the mixed norm sparse constraint in DBNs. We performed experiments using three different data sets for the task of hand written digit recognition and offered a practical approach for the use of overlapping groups with the mixed norm constraint. Although, our initial experiments were limited in the use of equal size overlapping groups, one could easily extended to non-symmetric overlapping groups, using a similar methodology. Inducing sparse constraints based on specific geometries may also provide better results in the case of digit recognition and offer more interesting results for tasks such as scene categorization.

\bibliography{xhalkias_icml12}
\bibliographystyle{plain}
\end{document}